\documentclass[conference]{IEEEtran}
\IEEEoverridecommandlockouts
\usepackage{cite}
\usepackage{amsmath,amssymb,amsfonts}
\usepackage{algorithmic}
\usepackage{graphicx}
\usepackage{textcomp}
\usepackage{xcolor}
\usepackage{bbding}
\usepackage{url}

\usepackage{pgfplots} 
\usetikzlibrary{patterns}

\def\BibTeX{{\rm B\kern-.05em{\sc i\kern-.025em b}\kern-.08em
    T\kern-.1667em\lower.7ex\hbox{E}\kern-.125emX}}
\begin{document}

\title{Sentence-Level Relation Extraction via Contrastive Learning with Descriptive Relation Prompts\\
}

\author{Jiewen Zheng,\, Ze Chen\ \\
         Interactive Entertainment Group of Netease Inc., Guangzhou, China \\
         \texttt{\{zhengjiewen02,jackchen\}@corp.netease.com}}

\maketitle

\begin{abstract}
Sentence-level relation extraction aims to identify the relation between two entities for a given sentence.
The existing works mostly focus on obtaining a better entity representation and adopting a multi-label classifier for relation extraction. 
A major limitation of these works is that they ignore background relational knowledge and the interrelation between entity types and candidate relations. 
In this work, we propose a new paradigm, Contrastive Learning with Descriptive Relation Prompts(CTL-DRP), to jointly consider entity information, relational knowledge and entity type restrictions. 
In particular, we introduce an improved entity marker and descriptive relation prompts when generating contextual embedding, and utilize contrastive learning to rank the restricted candidate relations.
The CTL-DRP obtains a competitive F1-score of 76.7\% on TACRED. Furthermore, the new presented paradigm achieves F1-scores of 85.8\% and 91.6\% on TACREV and Re-TACRED respectively, which are both the state-of-the-art performance.
\end{abstract}


\section{Introduction}
Relation Extraction(RE) is one of the fundamental information extraction tasks, aiming to obtain relational facts from given texts. 
Due to the capacity of extracting structured knowledge from unstructured texts, RE benefits many downstream applications, such as knowledge graph construction \cite{yu2021domain} and question answering \cite{dong2015question}.

Recently, the majority of methods regard sentence-level RE task as an entity pair classification problem \cite{zhong-chen-2021-frustratingly, ye2021pack}. 
Some of them focus on incorporating neural model with external information. 
Specially, \cite{peters2019knowledge} introduces entity embedding pre-trained from knowledge graph(KG) and \cite{zhang2019long} employs GCN\cite{kipf2016semi} to learn explicit relational knowledge from KG. 
Some others focus on extracting better entity representations from pre-trained language models(PLMs), e.g.\cite{zhou2021improved} and \cite{ye2021pack} employ different entity markers to make full use of names, spans, and types of both subject and object. 
More recently, \cite{lyu2021relation} takes mutual restrictions between relations and entity types into consideration, and builds different classifiers for different pairs of entity types.

However, we cannot well incorporate the external knowledge without a pre-built KG for a specific dataset. Meanwhile, most PLMs-based methods \cite{soares2019matching,joshi2020spanbert} treat relations as labels to be classified, and focus only on extracting entity representation for sentence embedding. 
Thus these methods inevitably ignore the background relational knowledge and neglect the potential semantics of relations.

To make the most of related information of entities and relations and the interrelations between entity types and relations, a new paradigm, Contrastive Learning with Descriptive Relation Prompts(CTL-DRP) is proposed. 
The paradigm exploits an improved entity marker with descriptive relation prompts to generate a sound entity-relation-aware contextual embedding, and employs contrastive learning to train a unified rank model rather than multiple classifiers \cite{lyu2021relation} when considering the restriction of entity types. The main contributions of this paper are summarized as follows:
\begin{itemize}
\item We propose CTL-DRP, a new paradigm specifically designed for relation extraction tasks. The CTL-DRP exploits entity types to generate candidate relations, and is trained via contrastive learning to ensure that the probability of a positive relation is larger than others.
\item We introduce an entity-relation-aware contextual embedding mechanism, which consists of an improved entity marker and descriptive relation prompts. The proposed mechanism can help integrate entity information and relation knowledge into the input text representation.
\item We evaluate our system on three different versions of RE benchmarks. The CTL-DRP achieves comparable results on TACRED \cite{zhang2017position} and outperforms previous SOTA methods on both TACREV \cite{alt-etal-2020-tacred} and Re-TACRED \cite{stoica2021re}.
\end{itemize}

\section{Methodology}
In this section, we first give an overview of the proposed paradigm in \ref{arch}, then we introduce the key parts of generating entity-relation aware embedding: improved entity marker and descriptive relation prompts in \ref{EntRep}. And in the following part \ref{ctl}, we give the details of contrastive learning with consideration of entity type restrictions. 

\subsection{Framework}\label{arch}
For sentence-level RE, given a sentence $x$ with a subject entity $e_s$ and an object entity $e_o$, the task is to predict the relation $r$ between $e_s$ and $e_o$ from $\mathcal{R}\cup$\{NA\}, where $\mathcal{R}$ is the pre-defined set of relations and NA denotes for no relation. The entity types of $e_s$ and $e_o$ are defined as $t_s$ and $t_o$.

\begin{figure*}[htbp]
\centering
\includegraphics[width=\textwidth]{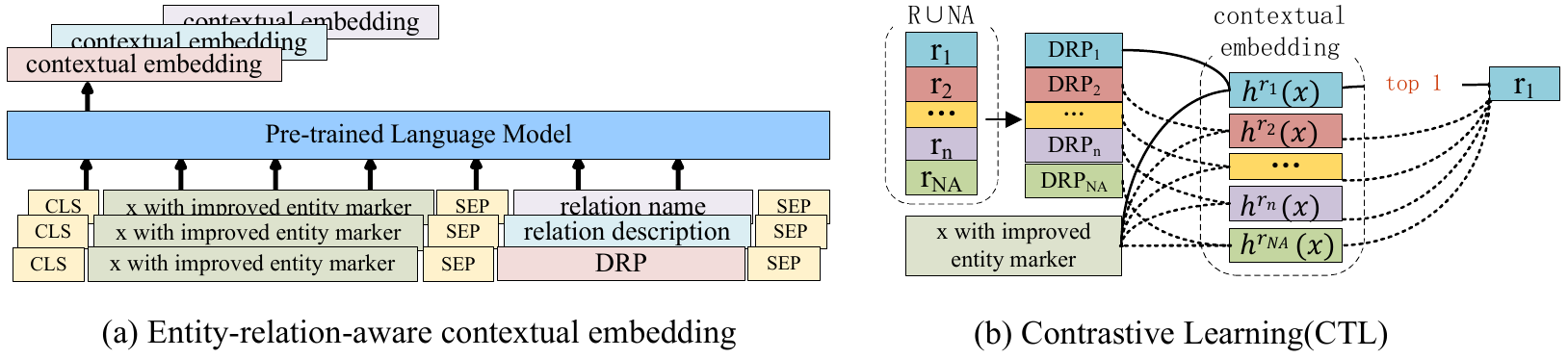}
\caption{(a) Pre-trained model computes an entity-relation-aware contextual embedding by feeding sentence $x$ with improved entity marker plus relation name/relation description/DRP. (b) $R$ is the set of candidate relation types for $x$. Each $r_i$ in $R$ corresponds to a $DPR_i$. Use contrastive learning to rank $r_i$ based on contextual embeddings. The true label in this example is $r_1$.}
\label{fig1}
\end{figure*}

The framework of the CTL-DRP, which consists of two modules: \textit{contextual embedding} and \textit{contrastive learning}, is shown in Fig. \ref{fig1}. For a given sentence $x$ with $e_s$ and $e_o$, we first utilize entity types to generate a restricted candidate relations set $R(t_s, t_o), R(t_s, t_o)\subset\mathcal{R}$. Then for each relation $r_i$ in $R(t_s, t_o)\cup\{NA\}$, a corresponding representation $h^{r_i}(x)$ for $x$ is generated by \textit{contextual embedding}. At last, the generated $h^{r_i}(x)$ is fed into \textit{contrastive learning}, and the relation in $R(t_s, t_o)\cup\{NA\}$ with the highest score is regarded as the final prediction.

\subsection{Contextual Embedding}\label{EntRep}
This module is initialized with a pre-trained language model. We take the embedding of the first subtoken from the last layer of the Transformer as the contextual embedding. To allowing the entity infomation and relational knowledge to be captured by the pre-trained language model simultaneously, we adopt an improved entity marker technique and descriptive relation prompts, which are presented in Sec. \ref{IEM} and Sec. \ref{DRP}.

\subsection{Improved Entity Marker}\label{IEM}
Composite entity information, such as names, spans, and NER types of both subject and object entities, provide useful clues to relation types \cite{zhou2021improved}. 
\cite{zhou2021improved} studies many existing entity representation techniques, and draws a conclusion that the typed entity marker outperforms others by a notable margin. 
We revisit the typed entity marker and slightly modify it to make a mutual promotion with DRP(Sec. \ref{DRP}). Differing from the typed entity marker, we explicitly express entity type and subject/object type information within parentheses added after entity names. 
The modified text is "[u1] SUBJ \emph{(subject subj-type)} [u2]......[u3]OBJ \emph{(object obj-type)}[u4]", where \emph{subj-type} and \emph{obj-type} are NER types of subjects and objects, [u1] to [u4] are new special tokens. 
An example of typed entity marker and improved entity marker is shown in Table \ref{tab5}.

\begin{table}[htbp]\scriptsize
\caption{Illustration of typed entity marker(punct) (TEM) and improved entity marker (IEM)}
\centering
\begin{tabular}{ll}
\hline
\hline
input:  & Sharpton is president of the National Action Network. \\
\hline          
TEM:  & \textcolor{red}{\# \^{} person \^{} Sharpton \#} is president of the \\
      & \textcolor{blue}{@ * organization * National Action Network @}.   \\
\hline 
IEM:  & \textcolor{red}{[u3] Sharpton (object person) [u4]} is president of the \\
      & \textcolor{blue}{[u1] National Action Network (subject organization) [u2]}.\\
\hline
\hline
\end{tabular}
\label{tab5}
\end{table}

\subsection{Descriptive Relation Prompts (DRP)}\label{DRP}
In order to incorporate contextual embeddings with relational knowledge, we first introduce manually written relation descriptions. 
Then we adopt narrative prompts, which indicate whether there is a relation between subject and object, to make a connection between entity representation and relation descriptions. 
The two main components of DRP, relation descriptions and narrative prompts, are described in detail below.

\begin{itemize}
\item \textbf{Relation Descriptions.} Normally a relation label, such as org:top-members/employees, is a abbreviation of the relationship holding between subject and object. 
We customize a description for each relation, which make pre-trained models capture semantic meaning of relation labels. 
The format \emph{(subject subj-type)} is exploited here to be consistent with the above improved entity marker. 
For example, the relation org:founded-by can be described as: \emph{subject organization is founded by object person}.  
See Appendix A for all relations on TACRED/Re-TACRED and their mapping descriptions.
\item \textbf{Prompts.} Furthermore, we use prompts as a connecting link to help model better understand the relation description.  
The prompt for a specific relation is: \emph{There is a relation between subject and object}. 
The prompt for NA is: \emph{There is no relation between subject and object}.
\item \textbf{DRP. } Thus, for a specific relation, the corresponding DRP is a prompt with the relation description followed by. 
For example, for \emph{org:founded-by}, its DRP is: \emph{There is a relation between subject and object: subject organization is founded by object person}. 
For no relation case, the descriptive relation prompt is the prompt itself. 
The example for DRP is in Table \ref{tab6}.
\end{itemize}

\begin{table}[htbp]\scriptsize
\caption{An example for relation description, prompt and DRP}
\centering
\begin{tabular}{ll}
\hline
\hline
relation name:  & org : top-members/employees \\
\hline
relation description: & object person is/are the top member /employees of\\
                      &   subject organization. \\   
\hline       
prompt:  & There is a relation between subject and object. \\
\hline
DRP:  & There is a relation between subject and object: \\
      & object person is/are the top member/employees of\\
      & subject organization.\\
\hline
\hline
\end{tabular}
\label{tab6}
\end{table}

\subsection{Contrastive Learning}\label{ctl}
The existing methods aim to choose the most likely relation from all given relations. However, entity types can restrict candidate relations and help us narrow down the scale of candidates. 
For example, the relation \emph{per:founded-by} is not likely to hold between \emph{person} and \emph{location}.

\begin{itemize}
\item \textbf{Candidate relations.} Candidate relations for a certain entity type pair $t_s$ and $t_o$ are denoted as $R(t_s, t_o)$. 
A function $H(r, t_s, t_o)$, which indicates whether the relation $r$ holds between subject type $t_s$ and object type $t_o$, is defined below:
\begin{equation}
H(r, t_s, t_o)= \begin{cases}
1& r \text{ } holds \text{ } between\text{ }  t_s \text{ } and \text{ } t_o\\
0& otherwise\\
\end{cases}\label{eq1}
\end{equation}

Where $r\in{\mathcal{R}}$ and $R(t_s, t_o)$ can be denoted as $R(t_s, t_o) = \{r | r\in{\mathcal{R}}, H(r,t_s,t_o)=1\}$.

For a sentence $x$ with subject type $t_s$ and object type $t_o$, the relation set $R(t_s, t_o)\cup{NA}$ includes all possible relation types for $x$. 
\item \textbf{CTL.} The size and elements of $R(t_s, t_o)$ differ from each other when considering different pairs of entity types. 
We utilize contrastive learning to generate final prediction from $R(t_s, t_o)\cup{NA}$ for a given sentence $x$, rather than multiple classifiers\cite{lyu2021relation} for each pair of entity types. 
For each relation $r_i$ in $R(t_s, t_o)\cup{NA}$, the corresponding contextual embedding of $x$ is denoted as $h^{r_i}(x)$. 
\begin{equation}
P(r_{i}|x) = \frac{exp(Wh^{r_i}(x) + b)}{\sum_{r_j\in{R(t_s, t_o)\cup{NA}}}exp(Wh^{r_j}(x) + b)}\label{eq2}
\end{equation} 

Where $P(r_{i}|x))$ is the probability of relation $r_i$ given the input text $x$, and  $W\in\mathbb{R}^d$, $b\in\mathbb{R}$ are model parameters. 
During training, the probability of the positive relation $r^+$ is maximized. 
\end{itemize}

\section{Experiments and Results}
We evaluate the CTL-DRP on three different versions of RE benchmarks: TACRED \cite{zhang2017position}, TACREV \cite{alt-etal-2020-tacred} and Re-TACRED \cite{stoica2021re}. By convention, we utilize micro-averaged F1-score to denote the performance of our paradigm.

\subsection{Dataset}
The statistics of TACRED, TACREV and Re-TACRED are listed in Table \ref{tab5}. 
For each dataset, each sample $S: = (x, e_s, t_s, e_o, t_o, r)$ includes a sentence $x$, a subject $e_s$ with type $t_s$, an object $e_o$ with type $t_o$, and a relation $r$ holding between $e_s$ and $e_o$. 

\begin{table}[htbp]\footnotesize
\caption{Statistics of datasets}
\centering
\begin{tabular}{ccccc}
\hline
\bf{Dataset}            & $\#$\textbf{train}     & $\#$\textbf{dev}     & $\#$\textbf{test}   & $\#$ \textbf{relations} \\
\hline
{\scriptsize TACRED}         & 68124  & 22631  & 15509  & 42  \\
{\scriptsize TACREV}         & 68124  & 22631  & 15509  & 42  \\
{\scriptsize Re-TACRED} & 58465     & 19584     & 13418 & 40  \\
\hline
\end{tabular}
\label{tab5}
\end{table}

\subsection{Experimental Setup}
Our experiment is implemented based on HuggingFace's Transformers\cite{wolf2020transformers}. 
The pre-trained models we adopt are \emph{sup-smi-roberta-large} and \emph{unsup-smi-roberta-large}\cite{gao2021simcse}. 
And we set batch size to 512 and learning rate to 1e-5, same as the batch size and learning rate used in pre-trained models' training. All experiments are trained in 10 epochs. We average the results of experiments on several random seeds.

\subsection{Experimental Results}
The extensive compared models for TACRED are: MTB \cite{soares2019matching}, KnowBert\cite{peters2019knowledge}, SpanBERT-ALT\cite{lyu2020auxiliary}, KEPLER\cite{wang2021kepler}, K-Adapter\cite{wang2020k}, LUKE\cite{yamada2020luke}, Typed maker\cite{zhou2021improved}, RECENT\cite{lyu2021relation} and DEEPSTRUCT\cite{wang-etal-2022-deepstruct}. The experimental results are shown in Table \ref{tab1}. 
CTL-DRP outperforms most of compared models, and achieves a comparable F1 score of 76.7\% with  DEEPSTRUCT\cite{wang-etal-2022-deepstruct}, which is the newest SOTA result. However, DEEPSTRUCT is a 10B parameter language
model whose model size is much larger than ours, and external datasets are utilized in DEEPSTRUCT. RECENT\cite{lyu2021relation} uses 14 classifiers while our CTL-DRP uses only one. 
With these models, although RECENT achieves a high precision, it has very limited recall. Compared to those works adopted just one single model, CTL-DRP achieves a considerable improvement, and the improvement on F1 exceeds 2 points. Besides, CTL-DRP$_{spanbert}$ outperforms its base model SpanBERT by almost 5 points on F1, indicating that the proposed paradigm CTL-DRP is effective on this RE task.


The compared models for TACREV are: BERT$_{large}$ + typed entity marker, RoBERTa$_{large}$ + typed entity
marker (punct), whose experimental results are reported in \cite{zhou2021improved}. We present our experimental results in  Table \ref{tab6}. CTL-DRP$_{sup}$ outperforms all the compared models, and achieves a new SOTA F1 score of 85.8\%.

The compared models for Re-TACRED are:  SpanBERT-large\cite{joshi2020spanbert}, Typed maker\cite{zhou2021improved} and Curriculum learning\cite{park2021improving}. 
The results are shown in Table \ref{tab2}. 
CTL-DRP also achieves a new state-of-the-art F1-score of 91.6\% on Re-TACRED.

\begin{table}[htbp]
\caption{F1-scores of compared models on TACRED test set}
\centering
\begin{tabular}{llll}
\hline
\bf{Model}            & \bf{P}     & \bf{R}     & \bf{F1}    \\
\hline
{\scriptsize MTB\cite{soares2019matching} }              & -     & -     & 71.5  \\
{\scriptsize KnowBert\cite{peters2019knowledge} }        & 71.6  & 71.6  & 71.6  \\
{\scriptsize SpanBERT\cite{joshi2020spanbert}}         & 70.8  & 70.9  & 70.8  \\
{\scriptsize SpanBERT-ALT\cite{lyu2020auxiliary}}     & 69.0  & 73.0  & 70.9  \\
{\scriptsize K-Adapter\cite{wang2020k}}        & 70.14 & 74.04 & 72.04 \\
{\scriptsize LUKE\cite{yamada2020luke}}             & 70.4  & 75.1  & 72.7  \\
{\scriptsize KEPLER\cite{wang2021kepler}}           & 71.5  & 72.5  & 72.0  \\
{\scriptsize Typed maker\cite{zhou2021improved}} & -     & -     & 74.6  \\
{\scriptsize RECENT\cite{lyu2021relation}}  & \textbf{90.9}  & 64.2  & 75.2  \\
{\scriptsize DEEPSTRUCT\cite{wang-etal-2022-deepstruct}}  & -  & -  & \textbf{76.8}  \\
\hline
\hline
$\rm{CTL\mbox{-}DRP_{spanbert}}$      & 76.8   & 74.6  & 75.7 \\
$\rm{CTL\mbox{-}DRP_{unsup}}$          & 73.8  & 74.6  & 74.2 \\
$\rm{CTL\mbox{-}DRP_{sup}}$         & 75.0  & \textbf{78.4}  & \textbf{76.7} \\
\hline
\end{tabular}
\label{tab1}
\end{table}

\begin{table}[htbp]
\caption{F1-scores of compared models on TACREV test set}
\centering
\begin{tabular}{llll}
\hline
\bf{Model}            & \bf{P}     & \bf{R}     & \bf{F1}    \\
\hline
{\scriptsize BERT-base \(+\) typed entity marker\cite{zhou2021improved}}  & -  & -  & 79.3  \\
{\scriptsize BERT-large \(+\) typed entity marker\cite{zhou2021improved}}         & -  & -  & 81.3  \\
{\scriptsize RoBERTa-large \(+\) typed entity maker\cite{zhou2021improved}} & -     & -     & 83.2  \\

\hline
\hline
$\rm{CTL\mbox{-}DRP_{unsup}}$         & 79.0  & 87.4  & 83.0 \\
$\rm{CTL\mbox{-}DRP_{sup}}$          & \textbf{85.9}  & \textbf{85.7}  & \textbf{85.8} \\
\hline
\end{tabular}
\label{tab6}
\end{table}

\begin{table}[htbp]
\caption{F1-scores of compared models on Re-TACRED test set}
\centering
\begin{tabular}{llll}
\hline
\bf{Model}            & \bf{P}     & \bf{R}     & \bf{F1}    \\
\hline
{\scriptsize SpanBERT-large\cite{joshi2020spanbert}}         & 70.8  & 70.9  & 70.8  \\
{\scriptsize Typed maker\cite{zhou2021improved}} & -     & -     & 91.1  \\
{\scriptsize Curriculum Learning\cite{park2021improving}}  & 91.3  & 91.5  & 91.4  \\
\hline
\hline
$\rm{CTL\mbox{-}DRP_{unsup}}$          & 91.2  & 91.2  & 91.2 \\
$\rm{CTL\mbox{-}DRP_{sup}}$         & \textbf{91.3}  & \textbf{91.9}  & \textbf{91.6} \\
\hline
\end{tabular}
\label{tab2}
\end{table}



\subsection{Ablation Study}
In this section, we first conduct ablation studies to investigate the  contribution of different components to our RE model, where we adopt \emph{sup-smi-roberta-large} in the experiments. Then we experiment without CTL-DRP
used to investigate the contribution of this new presented paradigm.
\begin{itemize}
\item \textbf{Entity representation techniques.} We use typed entity marker(punct)\cite{zhou2021improved}(TEM) to replace improved entity marker(IEM). 
Compared with TEM, IEM uses new special tokens instead of \emph{"@"} and \emph{"\#"} to enclose subjects and objects, and adds explicit subject/object type information after entity names. 
As shown in the second line of Table \ref{tab4}, the result of TEM drops 1.1/0.2 points on F1, justifying the effectiveness of new special tokens and subject/object type information. 

\begin{table}[htbp]\scriptsize
\caption{Ablation results on TACRED, TACREV and Re-TACRED.}
\centering
\resizebox{0.48\textwidth}{!}{
\begin{tabular}{ccccccc}
\hline
\begin{tabular}[c]{@{}c@{}}\textbf{entity}\\\textbf{repr.}\end{tabular}            & \begin{tabular}[c]{@{}c@{}}\textbf{relation}\\\textbf{name}\end{tabular}     & \begin{tabular}[c]{@{}c@{}}\textbf{relation}\\\textbf{desc.}\end{tabular}  & \bf{prompts}     & \begin{tabular}[c]{@{}c@{}}\textbf{TACRED}\\\textbf{F1}\end{tabular}   &   \begin{tabular}[c]{@{}c@{}}\textbf{TACREV}\\\textbf{F1}\end{tabular} & \begin{tabular}[c]{@{}c@{}}\textbf{Re-TACRED}\\\textbf{F1}\end{tabular} \\
\hline
\hline
IEM      &\XSolidBrush   & \CheckmarkBold  & \CheckmarkBold & \textbf{76.7} & \textbf{85.8} & \textbf{91.6} \\
TEM         &\XSolidBrush   & \CheckmarkBold  & \CheckmarkBold & 75.1(-1.1) & 83.1(-2.7) &91.4(-0.2)\\
IEM  & \XSolidBrush     & \CheckmarkBold  & \XSolidBrush    & 75.5(-0.7) & 83.6(-2.2) & 89.7(-1.9) \\
IEM  & \CheckmarkBold  &\XSolidBrush    &\XSolidBrush   & 74.1(-2.1)  & 81.1(-4.7) & 89.0(-2.6)\\
\hline
\end{tabular}
}
\label{tab4}
\end{table}

\item \textbf{DRP.} DRP consists of two parts: prompts and relation descriptions. We first use relation description to replace DRP. 
As shown in the third line of Table \ref{tab4}, the results of relation description drop 0.7/1.9 points on F1, indicating the necessity of using prompts as connecting links.  
We then use relation name instead of relation description. As shown in the last line of Table \ref{tab4}, the results of relation name drop 1.4/0.7 points on F1 compared to that of relation description, and drop 2.1/2.6 to CTL-DRP, indicating the importance of relation knowledge. 
\end{itemize}

For further study, we remove the whole CTL-DRP paradigm and directly employ the pre-trained model \emph{sup-smi-roberta-large} to do relation classification (indicated as RC$_{sup}$ for short). The experimental results are presented in Table \ref{tab7}. We can find that using the CTL-DRP can improve the performance on all the three RE datasets by a large margin.
\begin{table}[htbp]
\caption{F1-scores of compared models on different test sets}
\centering
\begin{tabular}{lccc}
\hline
\bf{Model}            & \begin{tabular}[c]{@{}c@{}}\textbf{TACRED}\\\textbf{F1}\end{tabular}     & \begin{tabular}[c]{@{}c@{}}\textbf{TACREV}\\\textbf{F1}\end{tabular}    & \begin{tabular}[c]{@{}c@{}}\textbf{Re-TACRED}\\\textbf{F1}\end{tabular}   \\
\hline
$\rm{CTL\mbox{-}DRP_{sup}}$         & \textbf{76.7}  & \textbf{85.8}  & \textbf{91.6} \\
$RC_{sup}$ &68.2 &76.23 &82.3 \\
\hline
\end{tabular}
\label{tab7}
\end{table}

\subsection{Analysis on Relation Types}
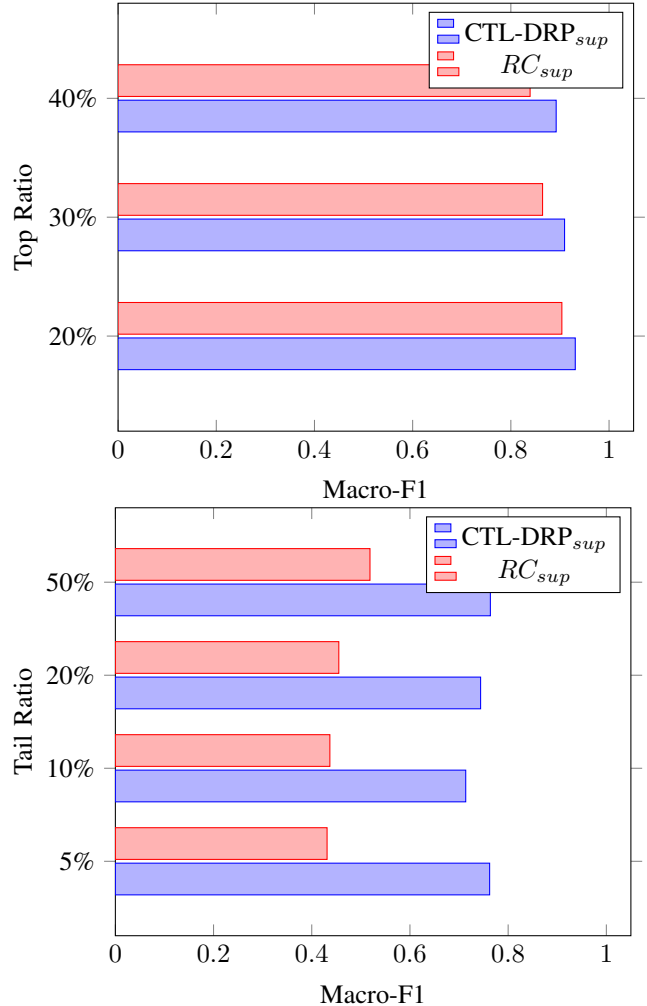
\begin{figure}
\begin{tikzpicture}
\begin{axis} [xbar = .05cm,
    bar width = 12pt,
    xmin = 0,
    xmax = 1.,
    ytick = data,
    yticklabels = {20\%, 30\%, 40\%},
    xlabel={Macro-F1},
    ylabel={Top Ratio},
    enlarge x limits = {value = .05, upper},
    enlarge y limits = {abs = .8}
]
 
\addplot coordinates {(0.9312,0) (0.9094,1) (0.8923,2)};
\addplot coordinates {(0.9039,0) (0.8646,1) (0.8392,2)};
 
\legend {CTL-DRP$_{sup}$, $RC_{sup}$};
 
\end{axis}
\end{tikzpicture}

\begin{tikzpicture}
\begin{axis} [xbar = .05cm,
    bar width = 12pt,
    xmin = 0,
    xmax = 1.,
    ytick = data,
    yticklabels = {5\%, 10\%, 20\%, 50\%},
    xlabel={Macro-F1},
    ylabel={Tail Ratio},
    enlarge x limits = {value = .05, upper},
    enlarge y limits = {abs = .8}
]
 
\addplot coordinates {(0.7623,0) (0.7136,1) (0.7439,2) (0.7638,3)};
\addplot coordinates {(0.4312,0) (0.4369,1) (0.4551,2) (0.5186,3)};
 
\legend {CTL-DRP$_{sup}$, $RC_{sup}$};
 
\end{axis}
\end{tikzpicture}

\caption{Macro-F1 score on RE-TACRED of different types of relations. The above figure indicates the top relations in train dataset and the below indicates the tail relations in train dataset.}
\label{fig2}
\end{figure}

To make clear how the CTL-DRP improves the performance on RE task, we investigate the performance of different types of relations on RE-TACRED. The macro-F1 score of different parts of relations are shown in Figure \ref{fig2}. We can find that relation classification model (RC$_{sup}$) can perform well on those relations with a high proportion in training sets, but performs poorly on the tail relations. However, CTL-DRP converts this RE classification task into a ranking task, it can improve the performance of relations with less data in the training sets. 

\section{Related Work}
\noindent\textbf{Sentence-level relation extraction.} One of the main ideas on sentence-level relation extraction is: combining PLMs with external knowledge. Some methods directly inject external knowledge to models in pre-training phase, like\cite{wang2020k}, \cite{yamada2020luke}. The other focus on continually pretraining PLMs via designed relation extraction related objectives, such as BERT-MTB\cite{soares2019matching}. Due to the cost of pretraining a language model, the idea of using information of relation extraction dataset rises. RECENT uses entity information to reduce the noisy candidate relation types and proves the effectiveness of using this kind information\cite{lyu2021relation}. We follow this direction and take a further step. We combine both the entity information and the label information into sentence-level embeddings, which enables the embedding to be entity-label aware.  

\noindent\textbf{Prompts.} Text classification tasks can be mapped into text entailment tasks\cite{yin2019benchmarking}. Specifically, a sentence s is treated as the premise. Using a prompt, such as "This news is about r", to combine with each class and generate a set of hypotheses(e.g. "This news is about sports.", "This news is about arts."). Present the entailment model with the premise s and the hypothesis set.  The class whose hypothesis gets the highest score from the model is chosen as the predicted label for s. We follow this method and use a hand-designed(hard prompt) prompt in our frame. In addition to hard prompt, prompt tuning is proved to be a simple and effective way for learning soft prompts, which can be tuned to incorporate signals from any number of labelled examples\cite{lester2021power}. 

\noindent\textbf{Contrastive learning objective.} Contrastive learning aims to pull "similar" data together while push "dissimilar" data apart\cite{hadsell2006dimensionality}. For a sample $x_i$, $x^{+}_{i}$ means its positive counterpart and $x^{-}_{i}$ means its negative counterpart. The model learns to make the representation of $x_i$ closer to its positive counterpart and away from the negative ones\cite{su2021tacl}. We follow the idea of contrastive learning object in\cite{chen2020simple}, \cite{karpukhin2020dense}, \cite{su2021tacl} and take a cross-entropy objective with in-batch negatives \cite{chen2017sampling}, \cite{henderson2017efficient}. For $x_i$, the batch is the hypothesis set defined in last paragraph. $x^{+}_{i}$ is the hypothesis with the true class label, and $x^{-}_{i}$ are the hypotheses with the noise class label. Let $h_i$ denotes the representation of $x_i$, the training objective for $x_i$ is:
\begin{equation}
l_i = -log\frac{exp(sim(h_i, h^{+}_i))}{\sum^{N}_{j=1}exp(sim(h_i, h_j))}\label{eq3}
\end{equation}
Where $N$ is the length of the hypotheses set of $x_i$, $h^{+}_i$ is the representation of positive hypothesis, and $h_j$ is the representation for both positive and negative hypotheses.

\section{Conclusion}

In this paper, we propose a novel paradigm CTL-DRP, the two key ideas are entity-relation-aware contextual embedding and contrastive learning. 
With improved entity marker, CTL-DRP integrates composite entity information. With relation descriptions, CTL-DRP captures semantic information of relation labels. 
With prompts, CTL-DRP adds connections between entities and descriptions. These features make original sentence embeddings become entity-relation-aware. 
Furthermore, CTL-DRP generates candidate relations, and ranks the relations based on contrastive learning in a unified model. CTL-DRP achieves new state-of-the-art F1-scores on both TACREV and Re-TACRED, and obtains a competitive F1-score of 76.7\% on TACRED which is 0.1\% less than the SOTA result. For the future work, we will investigate the performance of soft prompts and reduce prompts manually construction.

\bibliographystyle{IEEEtran}
\bibliography{re_20221228}

\begin{thebibliography}{10}
\providecommand{\url}[1]{#1}
\csname url@samestyle\endcsname
\providecommand{\newblock}{\relax}
\providecommand{\bibinfo}[2]{#2}
\providecommand{\BIBentrySTDinterwordspacing}{\spaceskip=0pt\relax}
\providecommand{\BIBentryALTinterwordstretchfactor}{4}
\providecommand{\BIBentryALTinterwordspacing}{\spaceskip=\fontdimen2\font plus
\BIBentryALTinterwordstretchfactor\fontdimen3\font minus
  \fontdimen4\font\relax}
\providecommand{\BIBforeignlanguage}[2]{{%
\expandafter\ifx\csname l@#1\endcsname\relax
\typeout{** WARNING: IEEEtran.bst: No hyphenation pattern has been}%
\typeout{** loaded for the language `#1'. Using the pattern for}%
\typeout{** the default language instead.}%
\else
\language=\csname l@#1\endcsname
\fi
#2}}
\providecommand{\BIBdecl}{\relax}
\BIBdecl

\bibitem{yu2021domain}
H.~Yu, H.~Li, D.~Mao, and Q.~Cai, ``A domain knowledge graph construction
  method based on wikipedia,'' \emph{Journal of Information Science}, vol.~47,
  no.~6, pp. 783--793, 2021.

\bibitem{dong2015question}
L.~Dong, F.~Wei, M.~Zhou, and K.~Xu, ``Question answering over freebase with
  multi-column convolutional neural networks,'' in \emph{Proceedings of the
  53rd Annual Meeting of the Association for Computational Linguistics and the
  7th International Joint Conference on Natural Language Processing (Volume 1:
  Long Papers)}, 2015, pp. 260--269.

\bibitem{zhong-chen-2021-frustratingly}
\BIBentryALTinterwordspacing
Z.~Zhong and D.~Chen, ``A frustratingly easy approach for entity and relation
  extraction,'' in \emph{Proceedings of the 2021 Conference of the North
  American Chapter of the Association for Computational Linguistics: Human
  Language Technologies}.\hskip 1em plus 0.5em minus 0.4em\relax Online:
  Association for Computational Linguistics, Jun. 2021, pp. 50--61. [Online].
  Available: \url{https://aclanthology.org/2021.naacl-main.5}
\BIBentrySTDinterwordspacing

\bibitem{ye2021pack}
D.~Ye, Y.~Lin, and M.~Sun, ``Pack together: Entity and relation extraction with
  levitated marker,'' \emph{arXiv preprint arXiv:2109.06067}, 2021.

\bibitem{peters2019knowledge}
M.~E. Peters, M.~Neumann, R.~L. Logan~IV, R.~Schwartz, V.~Joshi, S.~Singh, and
  N.~A. Smith, ``Knowledge enhanced contextual word representations,''
  \emph{arXiv preprint arXiv:1909.04164}, 2019.

\bibitem{zhang2019long}
N.~Zhang, S.~Deng, Z.~Sun, G.~Wang, X.~Chen, W.~Zhang, and H.~Chen, ``Long-tail
  relation extraction via knowledge graph embeddings and graph convolution
  networks,'' \emph{arXiv preprint arXiv:1903.01306}, 2019.

\bibitem{kipf2016semi}
T.~N. Kipf and M.~Welling, ``Semi-supervised classification with graph
  convolutional networks,'' \emph{arXiv preprint arXiv:1609.02907}, 2016.

\bibitem{zhou2021improved}
W.~Zhou and M.~Chen, ``An improved baseline for sentence-level relation
  extraction,'' \emph{arXiv preprint arXiv:2102.01373}, 2021.

\bibitem{lyu2021relation}
S.~Lyu and H.~Chen, ``Relation classification with entity type restriction,''
  \emph{arXiv preprint arXiv:2105.08393}, 2021.

\bibitem{soares2019matching}
L.~B. Soares, N.~FitzGerald, J.~Ling, and T.~Kwiatkowski, ``Matching the
  blanks: Distributional similarity for relation learning,'' \emph{arXiv
  preprint arXiv:1906.03158}, 2019.

\bibitem{joshi2020spanbert}
M.~Joshi, D.~Chen, Y.~Liu, D.~S. Weld, L.~Zettlemoyer, and O.~Levy, ``Spanbert:
  Improving pre-training by representing and predicting spans,''
  \emph{Transactions of the Association for Computational Linguistics}, vol.~8,
  pp. 64--77, 2020.

\bibitem{zhang2017position}
Y.~Zhang, V.~Zhong, D.~Chen, G.~Angeli, and C.~D. Manning, ``Position-aware
  attention and supervised data improve slot filling,'' in \emph{Conference on
  Empirical Methods in Natural Language Processing}, 2017.

\bibitem{alt-etal-2020-tacred}
\BIBentryALTinterwordspacing
C.~Alt, A.~Gabryszak, and L.~Hennig, ``{TACRED} revisited: A thorough
  evaluation of the {TACRED} relation extraction task,'' in \emph{Proceedings
  of the 58th Annual Meeting of the Association for Computational
  Linguistics}.\hskip 1em plus 0.5em minus 0.4em\relax Online: Association for
  Computational Linguistics, Jul. 2020, pp. 1558--1569. [Online]. Available:
  \url{https://aclanthology.org/2020.acl-main.142}
\BIBentrySTDinterwordspacing

\bibitem{stoica2021re}
G.~Stoica, E.~A. Platanios, and B.~P{\'o}czos, ``Re-tacred: Addressing
  shortcomings of the tacred dataset,'' in \emph{Proceedings of the
  Thirty-fifth AAAI Conference on Aritificial Intelligence}, 2021.

\bibitem{wolf2020transformers}
T.~Wolf, L.~Debut, V.~Sanh, J.~Chaumond, C.~Delangue, A.~Moi, P.~Cistac,
  T.~Rault, R.~Louf, M.~Funtowicz \emph{et~al.}, ``Transformers:
  State-of-the-art natural language processing,'' in \emph{Proceedings of the
  2020 conference on empirical methods in natural language processing: system
  demonstrations}, 2020, pp. 38--45.

\bibitem{gao2021simcse}
T.~Gao, X.~Yao, and D.~Chen, ``Simcse: Simple contrastive learning of sentence
  embeddings,'' \emph{arXiv preprint arXiv:2104.08821}, 2021.

\bibitem{lyu2020auxiliary}
S.~Lyu, J.~Cheng, X.~Wu, L.~Cui, H.~Chen, and C.~Miao, ``Auxiliary learning for
  relation extraction,'' \emph{IEEE Transactions on Emerging Topics in
  Computational Intelligence}, 2020.

\bibitem{wang2021kepler}
X.~Wang, T.~Gao, Z.~Zhu, Z.~Zhang, Z.~Liu, J.~Li, and J.~Tang, ``Kepler: A
  unified model for knowledge embedding and pre-trained language
  representation,'' \emph{Transactions of the Association for Computational
  Linguistics}, vol.~9, pp. 176--194, 2021.

\bibitem{wang2020k}
R.~Wang, D.~Tang, N.~Duan, Z.~Wei, X.~Huang, G.~Cao, D.~Jiang, M.~Zhou
  \emph{et~al.}, ``K-adapter: Infusing knowledge into pre-trained models with
  adapters,'' \emph{arXiv preprint arXiv:2002.01808}, 2020.

\bibitem{yamada2020luke}
I.~Yamada, A.~Asai, H.~Shindo, H.~Takeda, and Y.~Matsumoto, ``Luke: deep
  contextualized entity representations with entity-aware self-attention,''
  \emph{arXiv preprint arXiv:2010.01057}, 2020.

\bibitem{park2021improving}
S.~Park and H.~Kim, ``Improving sentence-level relation extraction through
  curriculum learning,'' \emph{arXiv preprint arXiv:2107.09332}, 2021.

\bibitem{wang-etal-2022-deepstruct}
\BIBentryALTinterwordspacing
C.~Wang, X.~Liu, Z.~Chen, H.~Hong, J.~Tang, and D.~Song, ``{D}eep{S}truct:
  Pretraining of language models for structure prediction,'' in \emph{Findings
  of the Association for Computational Linguistics: ACL 2022}.\hskip 1em plus
  0.5em minus 0.4em\relax Dublin, Ireland: Association for Computational
  Linguistics, May 2022, pp. 803--823. [Online]. Available:
  \url{https://aclanthology.org/2022.findings-acl.67}
\BIBentrySTDinterwordspacing

\bibitem{yin2019benchmarking}
W.~Yin, J.~Hay, and D.~Roth, ``Benchmarking zero-shot text classification:
  Datasets, evaluation and entailment approach,'' \emph{arXiv preprint
  arXiv:1909.00161}, 2019.

\bibitem{lester2021power}
B.~Lester, R.~Al-Rfou, and N.~Constant, ``The power of scale for
  parameter-efficient prompt tuning,'' \emph{arXiv preprint arXiv:2104.08691},
  2021.

\bibitem{hadsell2006dimensionality}
R.~Hadsell, S.~Chopra, and Y.~LeCun, ``Dimensionality reduction by learning an
  invariant mapping,'' in \emph{2006 IEEE Computer Society Conference on
  Computer Vision and Pattern Recognition (CVPR'06)}, vol.~2.\hskip 1em plus
  0.5em minus 0.4em\relax IEEE, 2006, pp. 1735--1742.

\bibitem{su2021tacl}
Y.~Su, F.~Liu, Z.~Meng, T.~Lan, L.~Shu, E.~Shareghi, and N.~Collier, ``Tacl:
  Improving bert pre-training with token-aware contrastive learning,''
  \emph{arXiv preprint arXiv:2111.04198}, 2021.

\bibitem{chen2020simple}
T.~Chen, S.~Kornblith, M.~Norouzi, and G.~Hinton, ``A simple framework for
  contrastive learning of visual representations,'' in \emph{International
  conference on machine learning}.\hskip 1em plus 0.5em minus 0.4em\relax PMLR,
  2020, pp. 1597--1607.

\bibitem{karpukhin2020dense}
V.~Karpukhin, B.~O{\u{g}}uz, S.~Min, P.~Lewis, L.~Wu, S.~Edunov, D.~Chen, and
  W.-t. Yih, ``Dense passage retrieval for open-domain question answering,''
  \emph{arXiv preprint arXiv:2004.04906}, 2020.

\bibitem{chen2017sampling}
T.~Chen, Y.~Sun, Y.~Shi, and L.~Hong, ``On sampling strategies for neural
  network-based collaborative filtering,'' in \emph{Proceedings of the 23rd ACM
  SIGKDD International Conference on Knowledge Discovery and Data Mining},
  2017, pp. 767--776.

\bibitem{henderson2017efficient}
M.~Henderson, R.~Al-Rfou, B.~Strope, Y.-H. Sung, L.~Luk{\'a}cs, R.~Guo,
  S.~Kumar, B.~Miklos, and R.~Kurzweil, ``Efficient natural language response
  suggestion for smart reply,'' \emph{arXiv preprint arXiv:1705.00652}, 2017.

\end{thebibliography}


\appendix

\section{Appendix}
\label{sec:appendix}
Table \ref{tablast1} presents our manually construction of relation descriptions.

\begin{table*}[]\footnotesize
\caption{Realtion Descriptions}
\begin{tabular}{|lp{13cm}|}
\hline
\multicolumn{2}{|c|}{Relation descriptions} \\ \hline
\multicolumn{1}{|l|}{org:founded-by}      & subject organization is founded by object person      \\ \hline
\multicolumn{1}{|l|}{org:top-members}      & object person is/are the top members/employees of subject organization      \\ \hline
\multicolumn{1}{|l|}{per:employee-of}      & subject person is/are employed by object organization      \\ \hline
\multicolumn{1}{|l|}{per:schools-attended}      & object school was where subject person attended      \\ \hline
\multicolumn{1}{|l|}{org:alternate-names}      & object name is an alias of subject organization                                                                                                              \\ \hline
\multicolumn{1}{|l|}{org:member-of}      & subject organization is a member of object organization/country/state/province      \\ \hline
\multicolumn{1}{|l|}{org:members}        & subject organization has object organization/country/state/province as one of its members \\ \hline
\multicolumn{1}{|l|}{org:parents}        & subject organization belongs to object organization/country/state/province \\ \hline
\multicolumn{1}{|l|}{org:subsidiaries}   & object organization/country/state/province belongs to subject organization \\ \hline
\multicolumn{1}{|l|}{org:shareholders}   & object organization/person is a shareholder of subject organization \\ 
\hline
\multicolumn{1}{|l|}{per:cities-of-residence}	& subject person lives in object city   \\ \hline
\multicolumn{1}{|l|}{per:city-of-death}	& subject person died in object city   \\ \hline
\multicolumn{1}{|l|}{per:city-of-birth} & subject person was born in object city   \\ \hline
\multicolumn{1}{|l|}{per:children}	& object person is a child of subject person   \\ \hline
\multicolumn{1}{|l|}{per:siblings}	& subject person and object person are siblings   \\ \hline
\multicolumn{1}{|l|}{per:spouse}	& subject person and object person are married   \\ \hline
\multicolumn{1}{|l|}{per:other-family}	& subject people come from different families   \\ \hline
\multicolumn{1}{|l|}{per:alternate-names}	& object name is an alias of subject person   \\ \hline
\multicolumn{1}{|l|}{per:parents}	& object person is parent of subject person   \\ \hline
\multicolumn{1}{|l|}{per:title}	& object designation is the job title of subject person   \\ \hline
\multicolumn{1}{|l|}{per:religion}	& object religious belief is the religion of subject person   \\ \hline
\multicolumn{1}{|l|}{per:age}	& object num is the age of subject person   \\ \hline
\multicolumn{1}{|l|}{org:website}	& object url link is the website of subject organization   \\ \hline
\multicolumn{1}{|l|}{per:stateorprovinces-of-residence}	& subject person lives in object state/province   \\ \hline
\multicolumn{1}{|l|}{per:stateorprovince-of-birth}	& subject person was born in object state/province   \\ \hline
\multicolumn{1}{|l|}{per:stateorprovince-of-death}	& subject person died in object state/province   \\ \hline
\multicolumn{1}{|l|}{per:countries-of-residence}	& subject person lives in object country   \\ \hline
\multicolumn{1}{|l|}{per:origin}	& subject person come from object country   \\ \hline
\multicolumn{1}{|l|}{per:country-of-birth}	& subject person was born in object country   \\ \hline
\multicolumn{1}{|l|}{per:country-of-death}	& subject person died in object country   \\ \hline
\multicolumn{1}{|l|}{org:city-of-headquarters}	& the head-quarter of subject organization locates in object city   \\ \hline
\multicolumn{1}{|l|}{org:country-of-headquarters}	& the head-quarter of subject organization locates in object country  \\ \hline
\multicolumn{1}{|l|}{org:stateorprovince-of-headquarters}	& the head-quarter of subject organization locates in object state/province  \\ \hline
\multicolumn{1}{|l|}{org:number-of-employees/members}	& object number is how many employees/members of subject organization  \\ \hline
\multicolumn{1}{|l|}{org:political/religious-affiliation}	& object political/religious-affiliation is political/religious orientation of subject organization  \\ \hline
\multicolumn{1}{|l|}{org:dissolved}	& object date was when subject organization was dissolved  \\ \hline
\multicolumn{1}{|l|}{org:founded}	& object date was when subject organization was founded  \\ \hline
\multicolumn{1}{|l|}{per:date-of-death}	& object date was when subject person died  \\ \hline
\multicolumn{1}{|l|}{per:date-of-birth}	& object date was when subject person was born  \\ \hline
\multicolumn{1}{|l|}{per:cause-of-death}	& object reason is what lead to subject person 's death  \\ \hline
\multicolumn{1}{|l|}{per:charges}	& object reason is why subject person was sentenced  \\ \hline
\multicolumn{1}{|l|}{org:city-of-branch}	& the branch of subject organization locates in object city  \\ \hline
\multicolumn{1}{|l|}{org:stateorprovince-of-branch}	& the branch of subject organization locates in object state/province  \\ \hline
\multicolumn{1}{|l|}{org:country-of-branch}	&the branch of subject organization locates in object country  \\ \hline
\multicolumn{1}{|l|}{per:identity}	& the subject person and the object person are the same person  \\ \hline 
\multicolumn{1}{|l|}{no-relation}	& there is no relation between subject and object  \\ \hline
\end{tabular}
\label{tablast1}
\end{table*}

\end{document}